\documentclass[11pt,a4paper]{article}

\PassOptionsToPackage{breaklinks}{hyperref}
\usepackage[hyperref]{acl2019}

\usepackage{times}
\usepackage[]{algorithm2e}
\usepackage[utf8]{inputenc} 
\usepackage[T1]{fontenc}    
\usepackage{booktabs}       
\usepackage{amsfonts}       
\usepackage{nicefrac}       
\usepackage{microtype}      
\usepackage{multirow}
\usepackage{caption}
\usepackage{amsmath,amssymb}
\usepackage{dsfont}			
\usepackage{cases}
\usepackage{color}
\usepackage{tikz}
\usetikzlibrary{shapes,arrows,calc}
\usepackage{caption}
\usepackage{textcomp}
\usepackage{subcaption}
\usepackage{enumitem}
\usepackage{import}
\usepackage{CJKutf8}
\usepackage[toc,page]{appendix}
\usepackage[textsize=large]{todonotes}
\usepackage[normalem]{ulem}
\useunder{\uline}{\ul}{}

\renewcommand{\vec}[1]{{\boldsymbol{\mathbf{#1}}}}
\DeclareMathOperator*{\argmax}{arg\,max}

\setlist[itemize]{leftmargin=15pt}

\newtheorem{defi}{Definition}

\aclfinalcopy 

\title{Saliency-driven Word Alignment Interpretation for\\ Neural Machine Translation}

\author{Shuoyang Ding\quad Hainan Xu\quad Philipp Koehn\\
   Center for Language and Speech Processing\\
   Johns Hopkins University\\
   {\tt \{dings, hxu31, phi\}@jhu.edu}}

\date{}

\begin{document}
\maketitle
\begin{abstract}
  Despite their original goal to jointly learn to align and translate, Neural Machine Translation (NMT) models, especially Transformer, are often perceived as not learning interpretable word alignments.
  In this paper, we show that NMT models do learn interpretable word alignments, which could only be revealed with proper interpretation methods.
  We propose a series of such methods that are model-agnostic, are able to be applied either offline or online, and do not require parameter update or architectural change.
  We show that under the force decoding setup, the alignments induced by our interpretation method are of better quality than fast-align for some systems, and when performing free decoding, they agree well with the alignments induced by automatic alignment tools. 
\end{abstract}

\section{Introduction}

Neural Machine Translation (NMT) has made lots of advancements since its inception.
One of the key innovations that led to the largest improvements is the introduction of the attention mechanism \cite{DBLP:journals/corr/BahdanauCB14,DBLP:conf/emnlp/LuongPM15},
which jointly learns word alignment and translation.
Since then, the attention mechanism has gradually become a general technique in various NLP tasks, including summarization \cite{DBLP:conf/emnlp/RushCW15,DBLP:conf/acl/SeeLM17}, natural language inference \cite{DBLP:conf/emnlp/ParikhT0U16} and speech recognition \cite{DBLP:conf/nips/ChorowskiBSCB15,DBLP:conf/icassp/ChanJLV16}.

\begin{figure}[h]
    \hspace{-0.5cm}
    \includegraphics[scale=0.85]{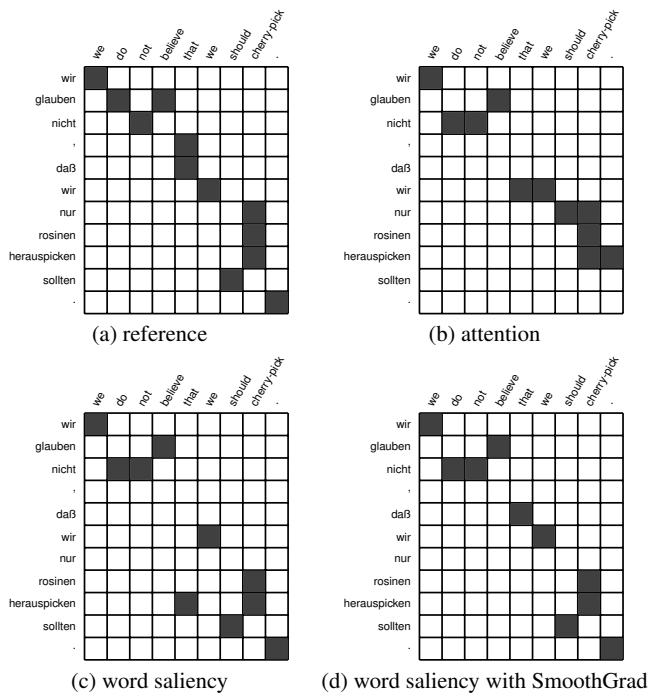}
    \caption{Comparison of our saliency-based word alignment interpretation of convolutional NMT model with reference and attention interpretation.}
    \label{fig:showcase}
\end{figure}


Although word alignment is no longer a integral step like the case for Statistical Machine Translation (SMT) systems \cite{DBLP:journals/coling/BrownPPM94,DBLP:conf/naacl/KoehnOM03}, there is a resurgence of interest in the community to study word alignment for NMT models.
Even for NMT, word alignments are useful for error analysis, inserting external vocabularies, and providing guidance for human translators in computer-aided translation.
When aiming for the most accurate alignments, the state-of-the-art tools include GIZA++ \cite{DBLP:journals/coling/BrownPPM94,DBLP:journals/coling/OchN03} and fast-align \cite{DBLP:conf/naacl/DyerCS13}, which are all external models invented in SMT era and need to be run as a separate post-processing step after the full sentence translation is complete.
As a direct result, they are not suitable for analyzing the internal decision processes of the neural machine translation models.
Besides, these models are hard to apply in the online fashion, i.e.\ in the middle of left-to-right translation process, such as the scenario in certain constrained decoding algorithms \cite{DBLP:conf/naacl/HaslerGIB18} and in computer-aided translation \cite{DBLP:conf/eacl/2014,arcan2014enhancing}.


For these cases, the current common practice is to simply generate word alignments from attention weights between the encoder and decoder.
However, there are problems with this practice. \citet{DBLP:conf/aclnmt/KoehnK17} showed that attention-based word alignment interpretation may be subject to ``off-by-one'' errors. \citet{DBLP:journals/corr/Zenkel,DBLP:conf/wmt/TangSN18,DBLP:conf/emnlp/RaganatoT18} pointed out that the attention-induced alignment is particularly noisy with  Transformer models.
Because of this, some studies, such as \citet{DBLP:conf/naacl/NguyenC18,DBLP:journals/corr/Zenkel} proposed either to add extra modules to generate higher quality word alignments, or to use these modules to further improve the model performance or interpretability.

This paper is a step towards interpreting word alignments from NMT without relying on external models.
We argue that using only attention weights is insufficient for generating clean word alignment interpretations,
which we demonstrate both conceptually and empirically.
We propose to use the notion of \emph{saliency} to obtain word alignment interpretation of NMT predictions.
Different from previous alignment models, our proposal is a pure interpretation method and \emph{does not require any parameter update or architecture change}.
Nevertheless, we are able to reduce Alignment Error Rate (AER) by 10-20 points over the attention weight baseline under two evaluation settings we adopt (see Figure~\ref{fig:showcase} for an example), and beat fast-align \cite{DBLP:conf/naacl/DyerCS13} by as much as 8.7 points.
Not only have we proposed a superior model interpretation method, but our empirical results also uncover that, contrary to common beliefs, architectures such as convolutional sequence-to-sequence models \cite{DBLP:conf/icml/GehringAGYD17} have already implicitly learned highly interpretable word alignments, which sheds light on how future improvement should be made on these architectures.

\section{Related Work}

We start with work that combines word alignments with NMT.
Research in this area generally falls into one of three themes: (1) employing the notion of word alignments to interpret the prediction of NMT; (2) making use of word alignments to improve NMT performance; (3) making use of NMT to improve word alignments.
We mainly focus on related work in the first theme as this is the problem we are addressing in this work.
Then we briefly introduce work in the other themes that is relevant to our study.
We conclude by briefly summarizing related work to our proposed interpretation method.

For the attention in RNN-based sequence-to-sequence model, the first comprehensive analysis is conducted by \citet{DBLP:conf/ijcnlp/GhaderM17}.
They argued that the attention in such systems agree with word alignment to a certain extent by showing that the RNN-based system achieves comparable alignment error rate comparable to that of bi-directional GIZA++ with symmetrization.
However, they also point out that they are not exactly the same, as training the attention with alignments would occasionally cause the model to forget important information.
\citet{DBLP:conf/emnlp/LeeSK17} presented a toolkit that facilitates study for the attention in RNN-based models.

There is also a number of other studies that analyze the attention in Transformer models.
\citet{DBLP:conf/emnlp/TangMRS18, DBLP:conf/wmt/TangSN18} conducted targeted evaluation of neural machine translation models in two different evaluation tasks, namely subject-verb agreement and word sense disambiguation.
During the analysis, they noted that the pattern in Transformer model (what they refer to as \textit{advanced attention mechanism}) is very different from that of the attention in RNN-based architecture, in that a lot of the probability mass is focused on the last input token.
They did not dive deeper in this phenomenon in their analysis.
\citet{DBLP:conf/emnlp/RaganatoT18} performed a brief but more refined analysis on each attention head and each layer, where they noticed several different patterns inside the modules, and
concluded that Transformer tends to focus on local dependencies in lower layers but finds long dependencies on higher ones.

Beyond interpretation, in order to improve the translation of rare words, \citet{DBLP:conf/naacl/NguyenC18} introduced LexNet, a feed-forward neural network that directly predicts the target word from a weighted sum of the source embeddings, on top of an RNN-based Seq2Seq models.
Their goal was to improve translation output and hence they did not empirically show AER improvements on manually-aligned corpora.
There are also a few other studies that inject alignment supervision during NMT training \cite{DBLP:conf/emnlp/MiWI16,DBLP:conf/coling/LiuUFS16}.
In terms of improvements in word alignment quality, \citet{DBLP:conf/wmt/LegrandAC16,DBLP:conf/acl/WangZAGN18,DBLP:conf/wmt/AlkhouliBN18} proposed neural word alignment modules decoupled from NMT systems, while \citet{DBLP:journals/corr/Zenkel} introduced a separate module to extract alignment from NMT decoder states, with which they achieved comparable AER with fast-align with Transformer models.

The saliency method we propose in this work draws its inspiration from visual saliency proposed by \citet{DBLP:journals/corr/SimonyanVZ13,DBLP:journals/corr/SpringenbergDBR14,DBLP:journals/corr/SmilkovTKVW17}.
It should be noted that these methods were mostly applied to computer vision tasks.
To the best of our knowledge, \citet{DBLP:conf/naacl/LiCHJ16} presented the only work that directly employs saliency methods to interpret NLP models.
Most similar to our work in spirit, \citet{DBLP:conf/acl/DingLLS17} used Layer-wise Relevance Propagation (LRP; \citealt{bach2015pixel}), an interpretation method resembling saliency, to interpret the internal working mechanisms of RNN-based neural machine translation systems.
Although conceptually LRP is also a good fit for word alignment interpretation, we have some concerns with the mathematical soundness of LRP when applied to attention models. Our proposed method is also considerably more flexible and easier to implement than LRP.

\section{The Interpretation Problem}

Formally, by interpreting model prediction, we are referring to the following problem:
given a trained MT model and input tokens $\mathcal{S} = \{s_0, s_1, \ldots, s_{I-1}\}$, at a certain time step $j$ when the models predicts $t_j$, we want to know which source word in $\mathcal{S}$ ``contributed'' most to this prediction.
Note that the prediction $t_j$ might not be $\argmax_{t_j} p(t_j\mid \vec{t_{1:j-1}})$, as the locally optimal option may be pruned during beam search and not end up in the final translation.

Under this framework, we can see an important conceptual problem regarding interpreting attention weights as word alignment.
Suppose for the same source sentence, there are two alternative translations that diverge at target time step $j$, generating $t_j$ and $t_j'$ which respectively correspond to different source words.
Presumably, the source word that is aligned to $t_j$ and $t_j'$ should changed correspondingly.
However, this is not possible with the attention weight interpretation, because the attention weight is computed \emph{before} prediction of $t_j$ or $t_j'$.
With that, we argue that an ideal interpretation algorithm should be able to adapt the interpretation with the specified output label, regardless of whether it is the most likely label predicted by the model.

As a final note,
the term ``attention weights'' here refers to the weights of the attention between encoder and decoder (the ``encoder-decoder attention'' in \citet{DBLP:conf/nips/VaswaniSPUJGKP17}).
Specifically, they do not refer to the weight of \emph{self-attention} modules that only exist in the Transformer architecture, which do not establish alignment between the source and target words. 

\section{Method}

Our proposal is based on the notion of visual saliency \cite{DBLP:journals/corr/SimonyanVZ13} in computer vision.
In brief, the saliency of an input feature is defined by the partial gradient of the output score with regard to the input.
We propose to extend this idea to NMT by drawing analogy between input pixels and the embedding look-up operation.

\subsection{Visual Saliency}


Suppose we have an image classification example $(\vec{x_0}, y_0)$, with $y_0$ being a specific image class and $\vec{x_0}$ being an $\left|\mathcal{X}\right|$-dimensional vector.
Each entry of $\vec{x_0}$ is an input feature (i.e., a pixel) to the classifier.
Given the input $\vec{x_0}$, a trained classifier can generate a prediction score for class $y_0$, denoted as $p(y_0\mid \vec{x_0})$.
Consider the first-order Taylor expansion of a perturbed version of this score at the neighborhood of input $\vec{x_0}$:

\begin{align}
p(y_0\mid\vec{x})&\approx p(y_0\mid\vec{x_0}) + \dfrac{\partial p(y_0\mid\vec{x})}{\partial \vec{x}}\bigg|_{\vec{x_0}}\cdot(\vec{x} - \vec{x_0}) \label{fm:saliency-taylor}
\end{align}
This is essentially re-formulating the perturbed prediction score $p(y_0\mid\vec{x})$ as an affine approximation of the input features, while the ``contribution'' of each feature to the final prediction being the partial derivative of the prediction score with regard to the feature.
Assuming a feature that is deemed as salient for the local perturbation of the prediction score would also be globally salient, the saliency of an input feature is defined as follows:

\begin{defi}
Denoted as $\vec{\Psi}(\vec{x}, y)$, the \textit{saliency} of feature vector $\vec{x}$ with regard to output class $y$ is defined as $\dfrac{\partial p(y\mid\vec{x})}{\partial \vec{x}}$. \label{def:saliency}
\end{defi} 
Note that $\vec{\Psi}(\vec{x}, y)$ is also a vector, with each entry corresponding to the saliency of a single input feature in $\vec{x}$. Such formulation has following nice properties:
\begin{itemize}[itemsep=-2pt,topsep=2pt]
\item The saliency of an input feature is related to the choice of output class $y$, as model scores of different output classes correspond to a different set of parameters, and hence resulting in different partial gradients for the input features.
This makes up for the aforementioned deficiency of attention weights in addressing the interpretation problem.
\item The partial gradient could be computed by back-propagation, which is efficiently implemented in most deep learning frameworks.
\item The formulation is agnostic to the model that generates $p(y\mid\vec{x})$, so it could be applied to any deep learning architecture.
\end{itemize}

\subsection{Word Saliency}

In computer vision, the input feature is a 3D Tensor corresponding to the level in each channel.
The key question to apply such method to NMT is what constitutes the input feature to a NMT system.
\citet{DBLP:conf/naacl/LiCHJ16} proposed to use the embedding of of the input words as the input feature to formulate saliency score, which results in the saliency of an input word being a vector of the same dimension as embedding vectors.
To obtain a scalar saliency value, they computed the mean of the absolute value of the embedding gradients.
We argue that there is a more mathematically principled way to approach this.

To start, we treat the word embedding look-up operation as a dot product between the embedding weight matrix $\vec{W}$ and an one-hot vector $\vec{z}$.
The size of $\vec{z}$ is the same as the source vocabulary size.
Similarly, the input sentence could be formulated as a matrix $\vec{Z}$ with only 0 and 1 entries.
Notice that $\vec{z}$ has certain resemblance to the pixels of an image, with each cell representing the pixel-wise activation level of the words in the vocabulary.
For the output word $t_j$ at time step $j$, we can similarly define the saliency of the one-hot vector $\vec{z}$ as:
\begin{align}
\vec{\Psi}(\vec{z}, t_j) = \dfrac{\partial p(t_j\mid\vec{Z})}{\partial \vec{z}}
\end{align}
where $p(t_j\mid\vec{Z})$ is the probability of word $t_j$ generated by the NMT model given source sentence $\vec{Z}$. $\vec{\Psi}(\vec{z}, t_j)$ is a vector of the same size as $\vec{z}$.

However, note that there is a key difference between $\vec{z}$ and pixels. If the pixel level is 0, it means that the pixel is black, while a 0-entry in $\vec{z}$ means that the input word is not the word denoted by the corresponding cell.
While the black region of an input image may still carry important information, we are not interested in the saliency of the 0-entries in $\vec{z}$.\footnote{Although we introduce $\vec{z}$ to facilitate presentation, note that word embedding look-up is never implemented as a matrix multiplication. Instead, it is implemented as a table look-up, so for each input word, only one row of the word embedding is fed into the subsequent computation. As a consequence, during training, since the other rows are not part of the computation graph, only parameters in the rows corresponding to the 1-entries will be updated. This is another reason why we choose to discard the saliency of 0-entries.}
Hence, we only take the 1-entries of matrix $\vec{Z}$ as the input to the NMT model.
For a source word $s_i$ in the source sentence, this means we only care about the saliency of the 1-entries, i.e., the entry corresponding to source word $s_i$:
\begin{align}
    \psi(s_i, t_j) &= \left[\dfrac{\partial p(t_j\mid\vec{Z})}{\partial \vec{z}}\right]_{s_i} \notag\\
    &= \left[\dfrac{\partial p(t_j\mid\vec{Z})}{\partial \vec{W_{s_i}}}\cdot \dfrac{\partial \vec{W_{s_i}}}{\partial\vec{z}}\right]_{s_i} \notag\\
    &= \left[\dfrac{\partial p(t_j\mid\vec{Z})}{\partial \vec{W_{s_i}}}\cdot \vec{W}\right]_{s_i} \notag\\
    &= \dfrac{\partial p(t_j\mid\vec{Z})}{\partial \vec{W_{s_i}}}\cdot \vec{W_{s_i}}
\end{align}
where $[\cdot]_i$ denotes the $i$-th row of a matrix or the $i$-th element of a vector.
In other words, the saliency $\psi(s_i, t_j)$ is a weighted sum of the word embedding of input word $s_i$, with the partial gradient of each cell as the weight. By comparison, the word saliency\footnote{\citet{DBLP:conf/naacl/LiCHJ16} mostly focused on studying saliency on the level of word embedding dimensions. This word-level formulation is proposed as part of the analysis in Section 5.2 and Section 6 of that work.} in \citet{DBLP:conf/naacl/LiCHJ16} is defined as:
\begin{align}
\psi'(s_i, t_j) = \text{mean}\left(\left|\dfrac{\partial p(t_j\mid\vec{Z})}{\partial \vec{W_{s_i}}}\right|\right)
\end{align}

There are two implementation details that we would like to call for the reader's attention:

\begin{itemize} \itemsep -2pt
\item When the same word occurs multiple times in the source sentence, multiple copies of embedding for such word need to be made to ensure that the gradients flowing to different instances of the same word are not merged;
\item Note that $\psi(s_i, t_j)$ is not a probability distribution, which does not affect word alignment results because we are taking $\argmax$. For visualizations presented herein, we normalized the distribution by $p(s_i\mid t_j) \propto \max(0, \psi(s_i, t_j))$. One may also use softmax function for applications that need more well-formed probability distribution.
\end{itemize}

\subsection{SmoothGrad\label{sec:smoothgrad}}

There are two scenarios where the naïve gradient-based saliency may make mistakes:
\begin{itemize} \itemsep -2pt
    \item For highly non-linear models, the saliency obtained from local perturbation may not be a good representation of the global saliency.
    \item If the model fits the distribution nearly perfectly, some data points or input features may become \textit{saturated}, i.e. having a partial gradient of 0.
    This does not necessarily mean they are not salient with regard to the prediction.
\end{itemize}

We alleviate these problems with SmoothGrad, a method proposed by \citet{DBLP:journals/corr/SmilkovTKVW17}.
The idea is to augment the input to the network into $n$ samples by adding random noise generated by normal distribution $\mathcal{N}(0, \sigma^2)$.
The saliency scores of each augmented sample are then averaged to cancel out the noise in the gradients.

We made one small modification to this method in our experiments: rather than adding noise to the word inputs that are represented as one-hot vectors, we instead add noise to the queried embedding vectors. This allows us to introduce more randomness for each word input.

\section{Experiments}

\subsection{Evaluation Method}

The best evaluation method would compare predicted word alignments against manually labeled word alignments between source sentences and NMT output sentences, but this is too costly for our study. Instead, we conduct two automatic evaluations for our proposed method using resources available:
\begin{itemize} \itemsep -2pt
  \item \emph{force decoding}: take a human-annotated corpus, run NMT models to force-generate the target side of the corpus and measure AER against the human alignment;
  \item \emph{free decoding}: take the NMT prediction, obtain reasonably clean reference alignments between the prediction and the source and measure AER against this reference.\footnote{Our reference alignment construction process is as follows: we first run automatic alignment on both sides, and take the intersection of the two outputs as ``sure'' alignments and the rest as ``possible'' alignments.}
\end{itemize}

Notice that both automatic evaluation methods have their respective limitation: the force decoding method may force the model to predict something it deems unlikely, and thus generating noisy alignment; whereas the free decoding method lacks authentic references.

\subsection{Setup\label{subsec:conf}}

We follow \citet{DBLP:journals/corr/Zenkel} in data setup and use the accompanied scripts of that paper\footnote{\tt https://github.com/lilt/alignment-scripts} for preprocessing.
Their training data consists of 1.9M, 1.1M and 0.4M sentence pairs for German-English (de-en), English-French (en-fr) and Romanian-English (ro-en) language pairs, respectively, whereas the manually-aligned test data contains 508, 447 and 248 sentence pairs for each language pair.
There is no development data provided in their setup, and it is not clear what they used for NMT system training, so we set aside the last 1,000 sentences of the training data for each language as the development set.

For our NMT systems, we use fairseq\footnote{\tt https://github.com/pytorch/fairseq} to train attention-based RNN systems (\textbf{LSTM}) \cite{DBLP:journals/corr/BahdanauCB14}, convolution systems (\textbf{FConv}) \cite{DBLP:conf/icml/GehringAGYD17}, and Transformer systems (\textbf{Transformer}) \cite{DBLP:conf/nips/VaswaniSPUJGKP17}. 
We use the pre-configured model architectures for IWSLT German-English experiments\footnote{The exact model options we used are respectively \texttt{fconv\_iwslt\_de\_en}, \texttt{lstm\_wiseman} \texttt{\_iwslt\_de\_en}, \texttt{transformer\_iwslt\_de\_en}.} to build all NMT systems.
Our experiments cover the following interpretation methods:
\begin{itemize} \itemsep -2pt
\item \emph{Attention}: directly take the attention weights as soft alignment scores. For transformer, we follow the implementation in fairseq and used the attention weights from the final layer averaged across all heads;
\item \emph{Smoothed Attention}: obtain multiple version of attention weights with the same data augmentation procedure as SmoothGrad and average them. This is to prove that smoothing itself does not improve the interpretation quality, and has to be used together with effective interpretation method;
\item \cite{DBLP:conf/naacl/LiCHJ16}: applied with normal back-propagation (\emph{Grad}) and \emph{SmoothGrad};
\item Ours: applied with normal back-propagation (\emph{Grad}) and \emph{SmoothGrad}.
\end{itemize}
For all the methods above, we follow the same procedure in \cite{DBLP:journals/corr/Zenkel} to convert soft alignment scores to hard alignment.

\begin{table*}[h]
\centering
\scalebox{0.85}{
\begin{tabular}{@{}lrrr|rrr|rrr@{}}
\toprule
 & \multicolumn{3}{c|}{\textbf{de\textless{}\textgreater{}en}} & \multicolumn{3}{c|}{\textbf{fr\textless{}\textgreater{}en}} & \multicolumn{3}{c}{\textbf{ro\textless{}\textgreater{}en}} \\
 & \multicolumn{1}{c}{\textbf{de-en}} & \multicolumn{1}{c}{\textbf{en-de}} & \multicolumn{1}{c|}{\textbf{bidir}} & \multicolumn{1}{c}{\textbf{en-fr}} & \multicolumn{1}{c}{\textbf{fr-en}} & \multicolumn{1}{c|}{\textbf{bidir}} & \multicolumn{1}{c}{\textbf{ro-en}} & \multicolumn{1}{c}{\textbf{en-ro}} & \multicolumn{1}{c}{\textbf{bidir}} \\ \midrule
\textbf{FConv} & \multicolumn{1}{l}{} & \multicolumn{1}{l}{} & \multicolumn{1}{l|}{} & \multicolumn{1}{l}{} & \multicolumn{1}{l}{} & \multicolumn{1}{l|}{} & \multicolumn{1}{l}{} & \multicolumn{1}{l}{} & \multicolumn{1}{l}{} \\
Attention & 38.5 & 40.1 & 37.5 & 23.8 & 27.4 & 22.0 & 40.9 & 38.6 & 39.1 \\
Smoothed Attention & 40.2 & 43.9 & 41.2 & 24.1 & 27.4 & 22.5 & 41.5 & 39.6 & 40.4 \\
\cite{DBLP:conf/naacl/LiCHJ16} Grad & 39.0 & 39.6 & 35.3 & 26.8 & 29.2 & 21.1 & 41.9 & 42.1 & 38.6 \\
\cite{DBLP:conf/naacl/LiCHJ16} SmoothGrad & 40.7 & 44.5 & 39.3 & 27.3 & 28.1 & 21.6 & 43.5 & 43.5 & 40.0 \\
Ours Grad & 33.1 & 40.5 & 26.8 & 25.2 & 22.7 & 11.9 & 37.1 & 39.4 & 29.8 \\
Ours SmoothGrad & {\ul \textbf{27.3}} & {\ul \textbf{33.0}} & {\ul \textbf{22.3}} & {\ul \textbf{21.2}} & {\ul \textbf{18.1}} & {\ul \textbf{8.5}} & {\ul \textbf{32.4}} & {\ul \textbf{34.2}} & {\ul \textbf{27.2}} \\
\textbf{LSTM} & \multicolumn{1}{l}{} & \multicolumn{1}{l}{} & \multicolumn{1}{l|}{} & \multicolumn{1}{l}{} & \multicolumn{1}{l}{} & \multicolumn{1}{l|}{} & \multicolumn{1}{l}{} & \multicolumn{1}{l}{} & \multicolumn{1}{l}{} \\
Attention & 42.8 & 47.5 & 36.9 & 33.7 & 38.0 & 25.8 & 47.1 & 47.0 & 40.9 \\
Smoothed Attention & 47.3 & 50.7 & 40.0 & 35.4 & 40.2 & 27.5 & 50.7 & 50.2 & 43.5 \\
\cite{DBLP:conf/naacl/LiCHJ16} Grad & 41.0 & 43.9 & 33.5 & 32.9 & 37.1 & 23.5 & 44.5 & 44.9 & 37.5 \\
\cite{DBLP:conf/naacl/LiCHJ16} SmoothGrad & 39.4 & 43.1 & 31.5 & 32.2 & 36.2 & 22.0 & 45.7 & 46.8 & 37.7 \\
Ours Grad & 47.5 & 50.2 & 38.6 & 41.1 & 41.6 & 30.4 & 54.2 & 55.8 & 42.8 \\
Ours SmoothGrad & {\ul 31.4} & {\ul 36.8} & {\ul 23.7} & {\ul 27.2} & {\ul 25.0} & {\ul 13.8} & {\ul 40.4} & {\ul 39.9} & {\ul 32.0} \\
\textbf{Transformer} & \multicolumn{1}{l}{} & \multicolumn{1}{l}{} & \multicolumn{1}{l|}{} & \multicolumn{1}{l}{} & \multicolumn{1}{l}{} & \multicolumn{1}{l|}{} & \multicolumn{1}{l}{} & \multicolumn{1}{l}{} & \multicolumn{1}{l}{} \\
Attention & 53.4 & 58.6 & 42.3 & 48.1 & 48.7 & 33.8 & 51.6 & 51.1 & 43.3 \\
Smoothed Attention & 55.8 & 56.1 & 48.6 & 42.5 & 47.5 & 32.9 & 57.5 & 57.6 & 51.5 \\
\cite{DBLP:conf/naacl/LiCHJ16} Grad & 51.1 & 56.2 & 43.7 & 43.6 & 47.9 & 39.9 & 46.7 & 48.4 & 35.5 \\
\cite{DBLP:conf/naacl/LiCHJ16} SmoothGrad & {\ul 36.4} & 45.8 & 30.3 & {\ul 27.0} & {\ul 25.5} & 15.6 & 41.3 & {\ul 39.9} & 33.7 \\
Ours Grad & 77.7 & 78.2 & 77.4 & 69.1 & 72.5 & 74.5 & 74.6 & 75.2 & 71.0 \\
Ours SmoothGrad & *{\ul 36.4} & {\ul 43.0} & *{\ul 29.0} & 29.7 & 25.9 & {\ul 15.3} & {\ul 41.2} & 41.4 & {\ul 32.7} \\ \midrule
fast-align Offline & 28.4 & 32.0 & 27.0 & 16.4 & 15.9 & 10.5 & 33.8 & 35.5 & 32.1 \\
fast-align Online & 30.8 & 34.4 & 30.0 & 18.8 & 16.8 & 13.6 & 37.1 & 41.1 & 35.9 \\
\cite{DBLP:journals/corr/Zenkel} & 26.6 & 30.4 & \textbf{21.2} & 23.8 & 20.5 & 10.0 & 32.3 & 34.8 & \textbf{27.6} \\
GIZA++ & \textbf{21.0} & \textbf{23.1} & 21.4 & \textbf{8.0} & \textbf{9.8} & \textbf{5.9} & \textbf{28.7} & \textbf{32.2} & 27.9 \\ \bottomrule
\end{tabular}
}
\caption{Alignment Error Rate (AER) with different saliency methods, under \emph{force decoding} setting. GIZA++ and fast-align Offline results are quoted from \citet{DBLP:journals/corr/Zenkel}, whereas fast-align Online stands for our online alignment result (c.f. Section \ref{subsec:conf}). \textit{bidir} refers to the symmetrized alignment results. Best results for each architecture are marked with underlines, and best interpretation/alignment results are respectively marked with boldface. Numbers affected by hyper-parameter tuning are marked with *.}
\label{tab:main-res1}
\end{table*}

For \emph{force decoding} experiments, we generate symmetrized alignment results with \texttt{grow-} \texttt{diag-final}.
We also include AER results\footnote{We reproduced the fast-align results as a sanity check and we were able to perfectly replicate their numbers with their released scripts.} of fast-align \cite{DBLP:conf/naacl/DyerCS13}, GIZA++\footnote{\tt https://github.com/moses-smt/giza-pp} and the best model (Add+SGD) from \citet{DBLP:journals/corr/Zenkel} on the same dataset for comparison.
However, the readers should be aware that there are certain caveats in this comparison:
\begin{itemize} \itemsep -2pt
\item All of these models are specifically designed and optimized to generate high-quality alignments, while our method is an \emph{interpretation} method and is not making any architecture modifications or parameter updates;
\item fast-align and GIZA++ usually need to update model with full sentence to generate optimal alignments, while our system and \citet{DBLP:journals/corr/Zenkel} can do so on-the-fly.
\end{itemize}

Realizing the second caveat, we also run fast-align under the \emph{online} alignment scenario, where we first train a fast-align model and decode on the test set.
This is a real-world scenario in applications such as computer-aided translation \cite{DBLP:conf/eacl/2014,arcan2014enhancing}, where we cannot practically update alignment models on-the-fly.
On the other hand, we believe this is a slightly better comparison for methods with online alignment capabilities such as \citet{DBLP:journals/corr/Zenkel} and this work.

The data used in \citet{DBLP:journals/corr/Zenkel} did not provide a manually-aligned development set, so we tune the SmoothGrad hyperparameters (noise standard deviation $\sigma$ and sample size $n$) on a 30-sentence subset of the German-English test data with the Transformer model.
We ended up using the recommended $\sigma=0.15$ in the original paper and a slightly smaller sample size $n=30$ for speed.
This hyperparameter setting is applied to the other SmoothGrad experiments as-is.
For comparison with previous work, we do not exclude these sentences from the reported results,
we instead mark the numbers affected to raise caution.

\subsection{Force Decoding Results}

\begin{table*}[h]
\centering
\scalebox{0.85}{
\begin{tabular}{@{}lrr|rr|rr@{}}
\toprule
 & \multicolumn{1}{c}{\textbf{de-en}} & \multicolumn{1}{c|}{\textbf{en-de}} & \multicolumn{1}{c}{\textbf{en-fr}} & \multicolumn{1}{c|}{\textbf{fr-en}} & \multicolumn{1}{c}{\textbf{ro-en}} & \multicolumn{1}{c}{\textbf{en-ro}} \\ \midrule
\textbf{FConv} & \multicolumn{1}{l}{} & \multicolumn{1}{l|}{} & \multicolumn{1}{l}{} & \multicolumn{1}{l|}{} & \multicolumn{1}{l}{} & \multicolumn{1}{l}{} \\
Attention & 27.4 & 24.2 & 20.7 & 23.6 & 32.5 & 25.6 \\
Smoothed Attention & 29.4 & 29.0 & 21.1 & 23.6 & 33.7 & 26.7 \\
\cite{DBLP:conf/naacl/LiCHJ16} Grad & 29.3 & 23.5 & 25.0 & 23.7 & 33.9 & 27.9 \\
\cite{DBLP:conf/naacl/LiCHJ16} SmoothGrad & 31.2 & 30.4 & 24.1 & 24.0 & 35.6 & 30.1 \\
Ours Grad & 18.2 & 20.0 & 20.2 & 14.3 & 24.9 & 22.8 \\
Ours SmoothGrad & {\ul \textbf{13.7}} & {\ul \textbf{14.2}} & {\ul \textbf{17.0}} & {\ul \textbf{10.6}} & {\ul \textbf{21.4}} & {\ul \textbf{17.4}} \\
\textbf{LSTM} & \multicolumn{1}{l}{} & \multicolumn{1}{l|}{} & \multicolumn{1}{l}{} & \multicolumn{1}{l|}{} & \multicolumn{1}{l}{} & \multicolumn{1}{l}{} \\
Attention & 33.6 & 34.6 & 32.5 & 32.3 & 36.5 & 31.7 \\
Smoothed Attention & 38.2 & 39.5 & 34.3 & 35.2 & 41.2 & 36.3 \\
\cite{DBLP:conf/naacl/LiCHJ16} Grad & 34.1 & 32.5 & 33.6 & 33.7 & 36.6 & 32.1 \\
\cite{DBLP:conf/naacl/LiCHJ16} SmoothGrad & 30.8 & 29.4 & 31.8 & 32.1 & 38.9 & 34.8 \\
Ours Grad & 35.9 & 36.7 & 40.2 & 36.3 & 44.1 & 43.1 \\
Ours SmoothGrad & {\ul 20.5} & {\ul 21.9} & {\ul 26.0} & {\ul 19.1} & {\ul 32.6} & {\ul 27.5} \\
\textbf{Transformer} & \multicolumn{1}{l}{} & \multicolumn{1}{l|}{} & \multicolumn{1}{l}{} & \multicolumn{1}{l|}{} & \multicolumn{1}{l}{} & \multicolumn{1}{l}{} \\
Attention & 50.2 & 53.0 & 50.4 & 48.5 & 44.9 & 41.9 \\
Smoothed Attention & 51.4 & 49.0 & 44.5 & 47.3 & 49.9 & 48.9 \\
\cite{DBLP:conf/naacl/LiCHJ16} Grad & 49.9 & 51.2 & 49.4 & 51.5 & 42.9 & 40.8 \\
\cite{DBLP:conf/naacl/LiCHJ16} SmoothGrad & 27.8 & 35.3 & {\ul 28.3} & 22.3 & 30.5 & {\ul 26.5} \\
Ours Grad & 76.7 & 76.6 & 77.1 & 78.9 & 71.9 & 74.0 \\
Ours SmoothGrad & *{\ul 26.6} & {\ul 31.0} & 30.0 & {\ul 21.4} & {\ul 30.0} & 28.2 \\ \bottomrule
\end{tabular}
}
\caption{Alignment Error Rate (AER) with different saliency models, under \emph{free decoding} setting. See the caption of Table \ref{tab:main-res1} for notations.}
\label{tab:main-res2}
\end{table*}

Table \ref{tab:main-res1} shows the AER results under the \emph{force decoding} setting.
First, note that after applying our saliency method with normal back-propagation, AER is only reduced for FConv model but instead increases for LSTM and Transformer.
The largest increase is observed for Transformer, where the AER increases by about 20 points on average. However, after applying SmoothGrad on top of that, we observe a sharp drop in AER, which ends up with 10-20 points lower than the attention weight baseline.
We can also see that this is not just an effect introduced by input noise, as the same smoothing procedure for attention increases the AER most of the times.
To summarize, at least under \emph{force decoding} settings, our saliency method with SmoothGrad obtains word alignment interpretations of much higher quality than the attention weight baseline.

As for \citet{DBLP:conf/naacl/LiCHJ16}, for FConv and LSTM architectures, it is not only consistently worse than our method, but at times also worse than attention.
Besides, the effect of SmoothGrad is also not as consistent on their saliency formulation as ours.
Although with the Transformer model, the \citet{DBLP:conf/naacl/LiCHJ16} method obtained better AER than our method under several settings, it is still pretty clear overall that the superior mathematical soundness of our method is translated into better interpretation quality.

While the GIZA++ model obtains the best alignment result in Table \ref{tab:main-res1}\footnote{While \citet{DBLP:conf/ijcnlp/GhaderM17} showed that the AER obtained by LSTM model is close to that of GIZA++, our experiments yield a much larger difference. We think this is largely due to the fact that we choose to train our model with BPE, while \citet{DBLP:conf/ijcnlp/GhaderM17} explicitly avoided doing so.}, most of our word alignment interpretation of FConv model with SmoothGrad surpasses the alignment quality of fast-align (either Online or Offline), sometimes by as much as 8.7 points (symmetrized ro\textless{}\textgreater{}en result).
Our best models are also largely on-par with \cite{DBLP:journals/corr/Zenkel}.
These are notable results as our method is an interpretation method and no extra parameter is updated to optimize the quality of alignment.
On the other hand, this also indicates that it is possible to induce high-quality alignments from NMT model without modifying its parameters, showing that it has acquired such information in an implicit way.
Most interestingly, although NMT is often deemed as performing poorly under low-resource setting, our interpretation seems to work relatively well on ro\textless{}\textgreater{}en language pair, which happens to be the language pair that we have least training data for.
We think this is a phenomenon that merits further exploration.

Besides, it can be seen that for all reported methods, the overall order for the number of alignment errors is FConv \textless{} LSTM \textless{} Transformer.
To our best knowledge, this is also a novel insight, as no one has analyzed attention weights of FConv with other architectures before.
We can also observe that while our method is not strong enough to fully bridge the gap of the attention noise level between different model architecture, it does manage to narrow the difference in some cases.

\subsection{Free Decoding Results}

Table \ref{tab:main-res2} shows the result under \emph{free decoding setting}. 
The trend in this group of experiment is similar to Table \ref{tab:main-res1}, except that Transformer occasionally outperforms LSTM.
We think this is mainly due to the fact that Transformer generates higher quality translations, but could also be partially attributed to the noise in fast-align reference.
Also, notice that the AER numbers are also generally lower compared to Table \ref{tab:main-res1} under this setting.
One reason is that our model is aligning output with which it is most confident, so less noise should be expected in the model behavior.
On the other hand, by qualitatively comparing the reference translation in the test set and the NMT output, we find that it is generally easier to align the translation as it is often a more literal translation.

\section{Analysis}


\subsection{Comparison with Li et al. (2016)}

\begin{figure}
    \centering
    \includegraphics[scale=0.81]{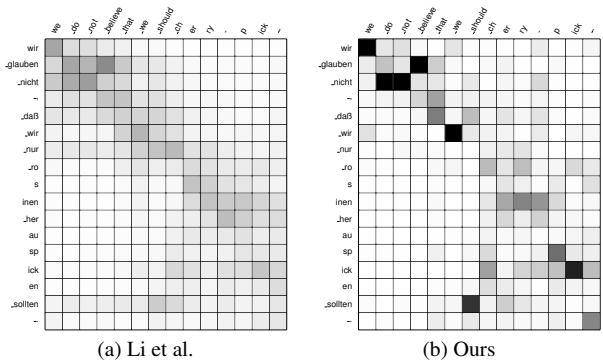}
    \caption{Saliency interpretation of FConv de-en model with the method in \citet{DBLP:conf/naacl/LiCHJ16} and this paper. SmoothGrad ($\sigma=0.15$, $n=30$) is applied for both interpretations.}
    \label{fig:li}
\end{figure}

The main reason why the word saliency formulation in \citet{DBLP:conf/naacl/LiCHJ16} does not work as well for word alignment is the lack of polarity in the formulation.
In other words, it only quantifies how much the input influences the output, but does not specify \textit{in what way} does the input influence.
This is sufficient for error analysis, but does not suit the purpose of word alignment, as humans will only align a target word to the input words that constitute a translation pair, i.e. have positive influence.

Figure \ref{fig:li} shows a case where this problem occurs in our German-English experiments. Note that in Subfigure (a), the source word \textit{nur} has high saliency on several target words, e.g. \textit{should}, but the word \textit{nur} is actually not translated in the reference.
On the other hand, as shown in Subfigure (b), our method correctly assigns negative (shown as white) or small positive values at all time steps for this source word.
Specifically, the saliency value of \textit{nur} for \textit{should} is negative with large magnitude, indicating significant negative contributions to the prediction of that target word.
Hence, a good word alignment interpretation should strongly avoid aligning them.

\subsection{SmoothGrad}

\begin{figure*}
    \centering
    \includegraphics[scale=0.81]{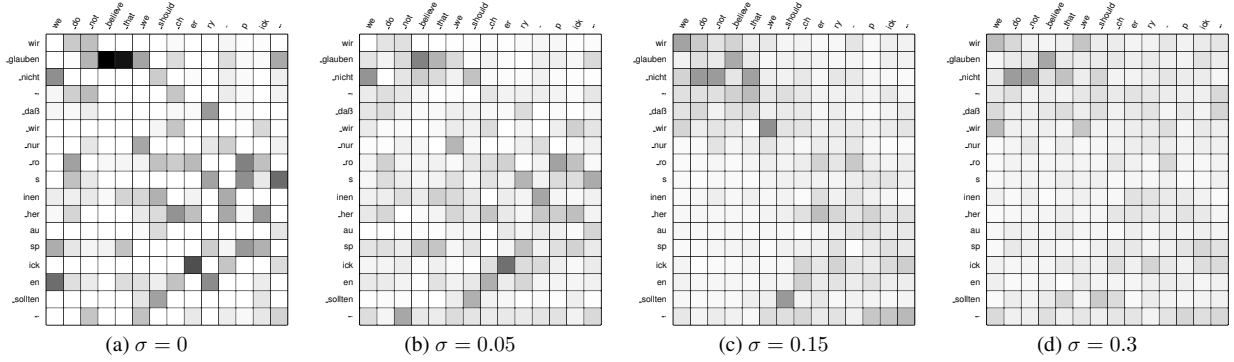}
    \caption{Saliency interpretation of Transformer de-en model with different SmoothGrad noise values $\sigma$ ($n=30$).}
    \label{fig:smoothgrad}
\end{figure*}

Tables \ref{tab:main-res1} and \ref{tab:main-res2} show that SmoothGrad is a crucial factor to reduce AER, especially for Transformer.
Figure~\ref{fig:smoothgrad} shows the interpretation of the same German-English sentence pair by our proposed method, but with Transformer and different SmoothGrad noise levels.
Specifically, Subfigures (a) and (c) corresponds to our Grad and SmoothGrad experiments in Table \ref{tab:main-res1}.
By comparing Subfigures (a) and (c), we notice that (1) without SmoothGrad, the word saliency obtained from the Transformer model is extremely noisy, and (2) the output of SmoothGrad is not only a smoother version of the na{\"i}ve gradient output,  but also gains new information by performing extra forward and backward evaluations with the noisy input. 
For example, compare the alignment point between source word \emph{wir} and target word \emph{we}: in Subfigure~(a), this word pair has very low saliency, but in (c), they become the most likely alignment pair for that target word.

Referring back to our motivation for using SmoothGrad in Section \ref{sec:smoothgrad}, we think the observations above verify that the Transformer model is a case where very high non-linearities occur almost everywhere in the parameter space, such that the saliency obtained from local perturbation is a very poor representation of the global saliency almost all the time.
On the other hand, this is also why the Transformer especially relies on SmoothGrad to work well, as the perturbation will give a better estimation of the global saliency.

It could also be observed from Subfigures~(b) and~(d) that when the noise is too moderate, the evaluation does not deviate enough from the original spot to gain non-local information, and at (d) it deviates too much and hence the resulting alignment is almost random. Intuitively, the noise parameter $\sigma$ should be sensitive to the model architecture or even specific input feature values, but interestingly we end up finding that a single choice  from the computer vision literature works well with all of our systems. We encourage future work to conduct more comprehensive analysis of the effect of SmoothGrad on more complicated architectures beyond convolutional neural nets.


\subsection{Alignment Dispersion}

\begin{table}[t]
\centering
\scalebox{0.8}{
\begin{tabular}{@{}llllll@{}}
\toprule
                     & \textbf{att} & \textbf{$\sigma$ = 0} & \textbf{$\sigma$ = 0.05} & \textbf{$\sigma$ = 0.15} & \textbf{$\sigma$ = 0.3} \\ \midrule
\textbf{FConv}       &              &                    &                       &                       &                      \\
force                & 2.09         & 1.36               & 1.48                  & 1.89                  & 2.59                 \\
free                 & 2.00         & 1.34               & 1.43                  & 1.79                  & 2.54                 \\ \midrule
\textbf{LSTM}        &              &                    &                       &                       &                      \\
force                & 1.75         & 1.63               & 2.02                  & 2.54                  & 2.89                 \\
free                 & 1.65         & 1.57               & 1.91                  & 2.46                  & 2.88                 \\ \midrule
\textbf{Transformer} &              &                    &                       &                       &                      \\
force                & 1.73         & 1.91               & 2.63                  & 2.76                  & 2.85                 \\
free                 & 1.69         & 1.89               & 2.62                  & 2.74                  & 2.84                 \\ \bottomrule
\end{tabular}
}
\caption{Alignment distribution entropy for selected de-en models. \textbf{att} stands for attention in Table \ref{tab:main-res1}.}
\label{tab:alg-dispersion}
\end{table}

We run German-English alignments under several different SmoothGrad noise deviation $\sigma$ and report their dispersion as measured by entropy of the (soft) alignment distribution averaged by number of target words.
Results are summarized in Table~\ref{tab:alg-dispersion}, where lower entropy indicates more peaky alignments.
First, we observe that dispersion of word saliency gets higher as we increase $\sigma$, which aligns with the observations in Figure~\ref{fig:smoothgrad}.
It should also be noted that the alignment dispersion is consistently lower for \emph{free decoding} than \emph{force decoding}.
This verifies our conjecture that the \emph{force decoding} setting might introduce more noise in the model behavior, but judging from this result, that gap seems to be minimal.
Comparing different architectures, the dispersion of attention weights does not correlate well with the dispersion of word saliency.
We also notice that, while the Transformer attention interpretation consistently results in higher AER, its dispersion is lower than the other architectures, indicating that
with attention, a lot of the probability mass might be concentrated in the wrong place more often.
This corroborates the finding in \citet{DBLP:conf/emnlp/RaganatoT18}.

\section{Discussion And Future Work}

There are several extensions to this work that we would like to discuss in this section.
First, in this paper we only explored two saliency methods among many others available \cite{DBLP:journals/dsp/MontavonSM18}.
In our preliminary study, we also experimented with guided back-propagation \cite{DBLP:journals/corr/SpringenbergDBR14}, a frequently used saliency method in computer vision, which did not work well for our problem.
We suspect that there is a gap between applying these methods on mostly-convolutional architectures in computer vision and architectures with more non-linearities in NLP. 
We hope the future research from the NLP and machine learning communities could bridge this gap.

Secondly, the alignment errors in our method comes from three different sources: the limitation of NMT models on learning word alignments, 
the limitation of interpretation method on recovering interpretable word alignments, and the ambiguity in word alignments itself.
Although we have shown that high quality alignment could be recovered from NMT systems (thus pushing our understanding on the limitation of NMT models), we are not yet able to separate these sources of errors in this work.
While exploration on this direction will help us better understand both NMT models and the capability of saliency methods in NLP, researchers may want to avoid using word alignment as a benchmark for saliency methods because of its ambiguity.
For such purpose, simpler tasks with clear ground truth, such as subject-verb agreement, might be a better choice.

Finally, as mentioned before, we are only conducting approximate evaluation to measure the ability of our interpretation method.
An immediate future work would be evaluating this on human-annotated translation outputs generated by the NMT system.

\section{Conclusion}

We propose to use word saliency and SmoothGrad to interpret word alignments from NMT predictions.
Our proposal is model-agnostic, is able to be applied either offline or online, and does not require any parameter updates or architectural change.
Both \emph{force decoding} and \emph{free decoding} evaluations show that our method is capable of generating word alignment interpretations of much higher quality compared to its attention-based counterpart.
Our empirical results also probe into the NMT black-box and reveal that even without any special architecture or training algorithm, some NMT models have already implicitly learned interpretable word alignments of comparable quality to fast-align. The model and code for our experiments are available at \texttt{https://github.com/shuoyangd/meerkat}.

\section*{Acknowledgements}
The authors would like to thank Matt Post for helpful feedback on an earlier draft of this work, and the authors of \citet{DBLP:journals/corr/Zenkel} for efforts in making their results easily reproducible. This material is based upon work supported in part by the DARPA LORELEI and IARPA MATERIAL programs.

\nocite{*}

\bibliography{acl2019}
\bibliographystyle{acl_natbib}


\end{document}